%% file: acl2021.tex
\documentclass[11pt,a4paper]{article}
\usepackage[hyperref]{acl2021}
\usepackage{times}
\usepackage{latexsym}

\usepackage{microtype}

\usepackage{stfloats }
\usepackage{multirow}
\usepackage{subfigure}
\usepackage{graphicx}
\usepackage{CJK}
\usepackage{tabularx}
\usepackage{longtable}
\usepackage[export]{adjustbox} 
\usepackage{colortbl}

\usepackage{newfloat}

\DeclareFloatingEnvironment{case}
\newcommand{\eop}{\textsc{Eop}}
\newcommand{\sep}{\textsc{Sep}}
\newcommand{\eos}{\textsc{Eos}}
\newcommand{\nl}{\textsc{nl}}
\title{Semantics of the Unwritten: The Effect of End of Paragraph and Sequence Tokens on Text Generation with GPT2}

\aclfinaltrue

\author{
He Bai,$^{1}$
Peng Shi,$^{1}$
Jimmy Lin,$^{1,2}$
Luchen Tan,$^{2}$
Kun Xiong,$^{2}$
Wen Gao,$^{4}$
Jie Liu,$^{3}$
Ming Li $^{1,2}$\\[0.1cm]
  $^{1}$ David R. Cheriton School of Computer Science, University of Waterloo \\
  $^{2}$ RSVP.ai  \quad  $^{3}$ Capital Normal University \\
  $^{4}$ School of Electronics Engineering and Computer Science, Peking University \\
  he.bai@uwaterloo.ca \\
  }

\date{}

\begin{document}
\maketitle

\input{sections/abstract.tex}
\input{sections/introduction.tex}
\input{sections/background.tex}
\input{sections/experiments.tex}

\input{sections/conclusion.tex}

\section*{Acknowledgments}
This work was supported by National Key Research and Development Program of China (2020AAA0109700), National Natural Science Foundation of China (62076167), and partially supported by NSERC OGP0046506.

We would like to thank Wei Zeng and his team in Peng Cheng Laboratory (PCL) for computing resources to support this project.

\bibliography{acl2021}
\bibliographystyle{acl_natbib}

\input{sections/appendix.tex}

\end{document}

%% file: sections/abstract.tex
\begin{abstract}
  The semantics of a text is manifested not only by what is read, but also by what is not read.
  In this article, we will study how the implicit ``not read'' information such as end-of-paragraph (\eop) and end-of-sequence (\eos) affect the quality of text generation. 
  Specifically, we find that the pre-trained language model GPT2 can generate better continuations by learning to generate the \eop~in the fine-tuning stage.
  Experimental results on English story generation show that \eop~can lead to higher BLEU score and lower \eos~perplexity. 
  We also conduct experiments on a self-collected Chinese essay dataset with Chinese-GPT2, a character level LM without \eop~or \eos~during pre-training.
  Experimental results show that the Chinese GPT2 can generate better essay endings with \eop.
  Our code is available on GitHub.\footnote{\url{https://github.com/rsvp-ai/semantic_unwritten}}
  \end{abstract}

%% file: sections/introduction.tex
\section{Introduction}


Large-pretrained neural models such as GPT~\cite{gpt} and BERT~\cite{DBLP:journals/corr/bert}  have achieved the state-of-the-art on many NLP tasks.
Among these models, OpenAI's GPT2~\cite{gpt2}, for example, has shown to be capable of generating long fluent text in many areas, such as stories~\cite{DBLP:conf/conll/SeePSYM19},  recipes~\cite{h2020recipegpt}, patent claims~\cite{DBLP:journals/corr/patentgpt}, and news~\cite{DBLP:conf/nips/grover}. 
However, the semantics of a text goes beyond what's written to what's not written:
When to break paragraphs and when to end. 
We wish to experiment on this issue: How much do \eop~and \eos~markers affect our ability to generate texts with GPT2.

To study the strength of GPT2 as a language generator, \citet{DBLP:conf/conll/SeePSYM19} conduct experiments in the context of story generation with the WritingPrompts~\cite{DBLP:conf/acl/LewisDF18} dataset. They find that the generated results of GPT2 have higher-quality content (using more rare words, concrete words, and named entities) by comparing the top 150 generated words. However, the average story length of the dataset is 12 paragraphs, 368 words.
In such lengthy human writings, the overall layout and text endings are also important, but whether the GPT2 can generate them properly is unclear, and how to generate better endings has not been investigated.


In this work, we find the generated endings are not only affected by \eos, but also \eop. 
\eop~can also help improve the topic relevance of the generated text.
We first conduct essay completion experiments with Chinese GPT2~\cite{GPT2-ML}, which is a character-level LM without \eos~or \eop~during pre-training.
Experimental results show that fine-tuning with \eop~can achieve higher ending quality score and topic relevance score in human evaluation.
We further conduct story generation experiments on dataset WritingPrompts with English GPT2-117, which holds the line break ``\textbackslash n''~(\nl) in the vocabulary.
Thus, the \nl~can be treated as the end-of-paragraph during fine-tuning~\cite{DBLP:conf/emnlp/MaoMMC19}.
Experimental results show that learning to end the paragraph can benefit the word/token perplexity, BLUE score, \eos~perplexity, and human evaluated ending quality score.

Our contributions are as follows: 
We show that not only the well-known \eos~but also the \eop, is part of the semantics of a text, and training with this information improves the text generation itself.
The paragraph information not only can help improve the effectiveness of the generation model but also help to generate the end of the text.
We also investigate different approaches to incorporating paragraph information into the LM generator.
Our findings indicate that \sep/\eop~and \eos~should be introduced to GPT2 types of models during pre-training, to generate better text in length.

%% file: sections/background.tex
\section{Background}

Our target task is to conduct auto-regressive language modeling over WritingPrompts and the ChineseEssay dataset.
The basic assumption of auto-regressive generation is that the probability of a word sequence equals the product of conditional word probability:
$P\left( w_{1:T}|W_{0}\right) =\prod_{t=1}^{T}P\left( w_{t}|w_{1:t-1},W_{0}\right)$
where $W_{0}$ is the given context, and in this work, $W_{0}$ can be a story prompt or the beginning of an essay.
The generated sequence length $T$ is usually determined by the time $t$ generating the \eos~(end-of-sequence) token:
$ P\left(w_{T}|w_{1:T-1}, W_{0}\right) =P\left(\eos|w_{1:t-1},W_{0}\right)$
In this work, the model computing the conditional probabilities is self-attention Transformer~\cite{DBLP:conf/nips/VaswaniSPUJGKP17}.
We train our model with the cross-entropy loss between the predicted conditional probabilities and the ground-truth next token.


When the target of generation consists of multiple paragraphs, there are several approaches to indicating the paragraph ending.
The most common and obvious approach is to separate paragraphs with line break \nl:
$ w_{1:T} = p_{1}, \nl, ..., p_{n-1}, \nl, p_{n}, \eos~$
where $p_i = \{w_{b_i:e_i}\}$ is the words sequence of paragraph $i$, from the beginning word $w_{b_i}$ to the ending word $w_{e_i}$.
However, 
not every paragraph ends with \nl, and during the pre-training, not every \nl~represents the paragraph separator~(\sep) .
A better option is to append a new specific token \eop~to indicate the end of the paragraph:
  $w_{1:T} = p_{1}^{,}, ..., p_{n-1}^{,}, p_{n}^{,}, \eos$
where $p_{i}^{,} = \{w_{b_i:e_i},\eop\}$.
Then, each paragraph can end with the \eop~and the transformer-based language model can learn this feature with every paragraph in the training data, without distinguishing when to generate \eop~and when not to.


It is well known that greedy decoding and beam search usually lead to repetitive and degenerate outputs\cite{DBLP:conf/acl/ShangLL15, DBLP:journals/corr/abs-1911-03587}. 
Sampling-based decoding methods have shown a strong ability in generating diversity, fluency and repetitiveness of the generation with pre-trained language models, such as \textit{top-k} and \textit{top-p} sampling.
In this work, we choose the \textit{top-p} sampling decoding algorithm and set the $p$ equals to 0.95.

%% file: sections/experiments.tex
\section{Experiments}

\subsection{Datasets}
\label{sec:datasets}
\begin{table}[t]
  \centering  \small
  \begin{tabular}{l|cc}
  Dataset &Story &Essay\\ 
  \hline
  Language&English &Chinese\\
  \hline
  \#Train samples &199,083 &1,615\\
  \#Test samples &11,069 &461\\
  \#Validation samples &11,474&195 \\
  \hline
  \#Avg. words per sample &367.9& 571.3\\
  \#Avg. paragraphs per sample &12.1&5.6 \\
  \hline
  \end{tabular}
  \caption{\label{table:datasets}Detailed information of the filtered WritingPrompts dataset and the ChineseEssay dataset.} 
  \end{table}

\smallskip \noindent \textbf{Story Generation.}
The story generation dataset is the WritingPrompts, collected by~\citet{DBLP:conf/acl/LewisDF18} from Reddit.
It is a large dataset of 300K human-written stories.
Each instance of this dataset is the pair of a short prompt and a long story.
Following~\citet{DBLP:conf/conll/SeePSYM19}, we exclude examples that are longer than 1024 BPE tokens to meet the maximum length restriction of GPT2.
Statistics for this dataset are detailed in Table~\ref{table:datasets}. We sample 1000 examples from the test set for decoding.

\smallskip \noindent \textbf{Essay Completion.}
We build an essay completion dataset ChineseEssay, which is collected from primary school and annotated by native Chinese annotators.
All these essays are descriptive essays about people, such as family members and teachers.
Hence, compared with the WritingPrompts, this dataset is smaller but less open domain.
Dataset statistics are also shown in Table~\ref{table:datasets}. 

\begin{table*}[ht!]
  \centering \small
  \begin{tabular}{l|c|c|c|c|c|c|c|c|c|c}
  \hline
  \textbf{ParaType}&\textbf{FT}&\textbf{\eos}&\textbf{T PPL}&\textbf{T PPL(-)}&\textbf{BLEU1}&\textbf{BLEU2}&\textbf{DIST1}&\textbf{DIST2}&\textbf{\eos\%}&\textbf{\eos~PPL}\\
  \hline
  \multirow{3}{*}{None} & No &No &12.12&11.48&33.6&7.5&34.46&73.95&0&-\\ 
  \cline{2-11}
   & Yes&No &11.44&11.44&38.1 &9.9 &32.95 & 73.96&0&-\\ 
   \cline{2-11}
   & Yes&Yes &10.43&\textbf{10.42}&42.7&10.7&37.57&78.26&76.41&22.15\\ 
  \hline
  \sep~DIY& Yes&Yes &10.45&10.52&44.1&11&38.73&78.98&90.26&8.92\\ 
  \hline
  \eop~DIY& Yes&Yes &\textbf{10.34}&10.48&\textbf{45.4}&\textbf{11.2}&\textbf{40.18}&\textbf{80.61}&\textbf{93.07}&\textbf{2.74}\\ 
  \hline
  \end{tabular}
  \caption{\label{table:results_essay}Test results of different models with/without fine-tuning(FT) on ChineseEssay dataset.} 
  \vspace{-2mm}
  \end{table*}

  \begin{table*}[t]
    \centering \small
    \begin{tabular}{l|c|c|c|c|c|c|c|c|c|c}
    \hline
    \textbf{ParaType}&\textbf{FT}&\textbf{W PPL}&\textbf{W PPL(-)}&\textbf{T PPL}&\textbf{T PPL(-)}&\textbf{\eos~PPL}&\textbf{BLEU1}&\textbf{BLEU2}&\textbf{DIST1}&\textbf{DIST2}\\
    \hline
    \multirow{2}{*}{None} & No &42.53&42.20&34.42&34.17&295.50&20.3&2.2&\textbf{58.87}&\textbf{89.78}\\ 
    \cline{2-11}
     & Yes &31.34&31.35&25.81&25.81&4.63&30.4&4.6&50.07&87.12\\ 
    \hline
    \multirow{2}{*}{\sep~}  & No &39.97&42.00&32.43&33.79&102.93&20.3&2.2&58.87&89.78\\ 
    \cline{2-11}
     & Yes &\textbf{29.36}&\textbf{31.24}&\textbf{24.23}&\textbf{25.57}&4.32&31.2&4.3&50.15&85.88\\ 
    \hline
    \sep~DIY& Yes &30.23&32.17&24.99&26.38&4.48&31.5&6.8&48.57&83.84\\ 
    \hline
    \multirow{2}{*}{\eop~\nl}& No &40.10&41.84&32.52&33.68&26478.91&20.3&2.2&58.87&89.78\\ 
    \cline{2-11}
     & Yes &29.95&31.32&24.70&25.63&20534.60&30.7&4.3&49.79&85.44\\ 
    \hline
    \eop~DIY& Yes &30.18&32.21&24.95&26.41&\textbf{2.26}&\textbf{31.7}&\textbf{6.9}&48.32&83.82\\ 
    \hline
    \end{tabular}
    \caption{\label{table:results_story}Test results on WritingPrompts dataset. } 
    \end{table*}

\subsection{Experimental Settings}
\label{sec:settings}
\smallskip \noindent \textbf{Model Configuration.}
For Chinese essay generation, we use Chinese-GPT2~\cite{GPT2-ML}, which is a 48 layers Transformer with 1.5 billion parameters, pre-trained with 15GB Chinese corpus.
For story generation, we fine-tune the OpenAI's GPT2-117 with WritingPrompts following previous work~\cite{DBLP:conf/conll/SeePSYM19, DBLP:conf/emnlp/MaoMMC19}.
The GPT2-117 model has 12 layers and 117 million parameters.
During fine-tuning, the batch size is 32 and the warm-up steps are 800. 
The other hyperparameters are the same as the default setting of Huggingface Transformers~\cite{DBLP:journals/corr/abs-1910-03771}.
Models can converge after 15 epochs for GPT2-117 and 3 epochs for Chinese-GPT2. 
The checkpoints with the best evaluation results on validation set are chosen for further testing.

\smallskip \noindent \textbf{Automatic Metrics.}
We use the following metrics: perplexity over all words/tokens~(W/T PPL); perplexity over words/tokens excluding \eos/\eop/\sep~(W/T PPL(-)); perplexity of \eos~(\eos~PPL); percentage of the generated texts that are ending with \eos~(\eos\%); BLEU/Distinct score excluding \eos/\eop/\sep~(BLEU/DIST). All perplexities are macro-average.

\smallskip \noindent \textbf{Human Evaluation Metrics.}
We also conduct the human evaluation with 50 random samples from the test set. 
For ChineseEssay, we collect generations from \eos~fine-tuned model, \eos+\eop~fine-tuned model, and \eos+\sep~fine-tuned model.
For WritingPrompts, we collect generations from the model fine-tuned with \eop~and the model without \sep/\eop. 
Four native speakers are asked to compare the generations of different systems in pairs over four metrics:\ topic relevance, fluency, ending quality, and overall preference. 
The assessors were presented with pairs of generated output and asked to make a three way judgment:\ whether the ``left system'' was better, the ``right system'' was better, or ``cannot distinguish''.
The latter option either meant that both output were equally good, or equally bad.
To prevent inadvertent bias, all systems were blinded, i.e., the assessors did not know which system generated which output, and presentation order was randomized.
After annotation, we count the total times of each system outperforming the others, and then normalize to 0-100\%.
\begin{figure*}[h]
  \centering \small
    \subfigure[Global View]
    {\includegraphics[width=0.40\textwidth]{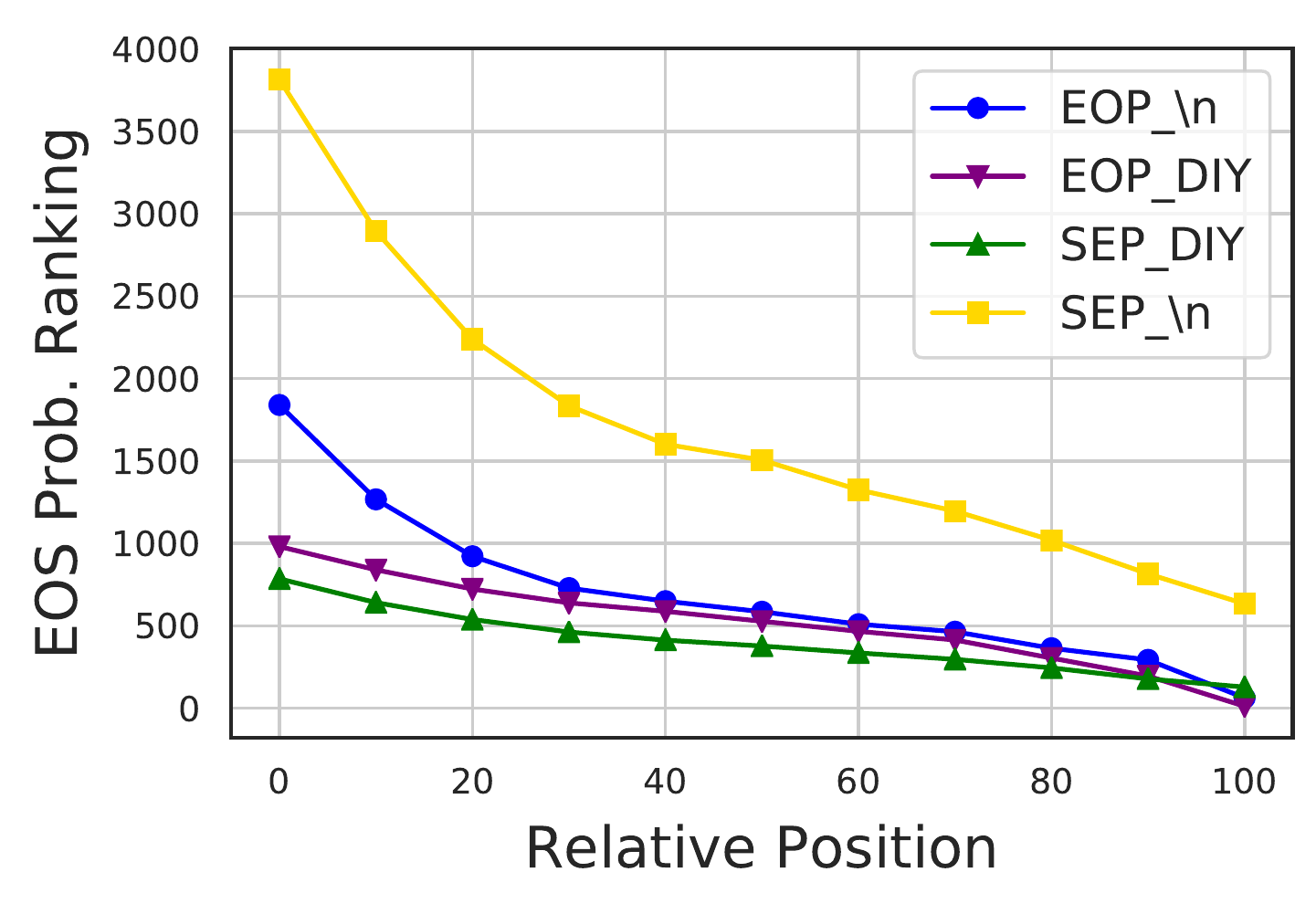}\label{fig:global}}
    \subfigure[Local View]
    {\includegraphics[width=0.40\textwidth]{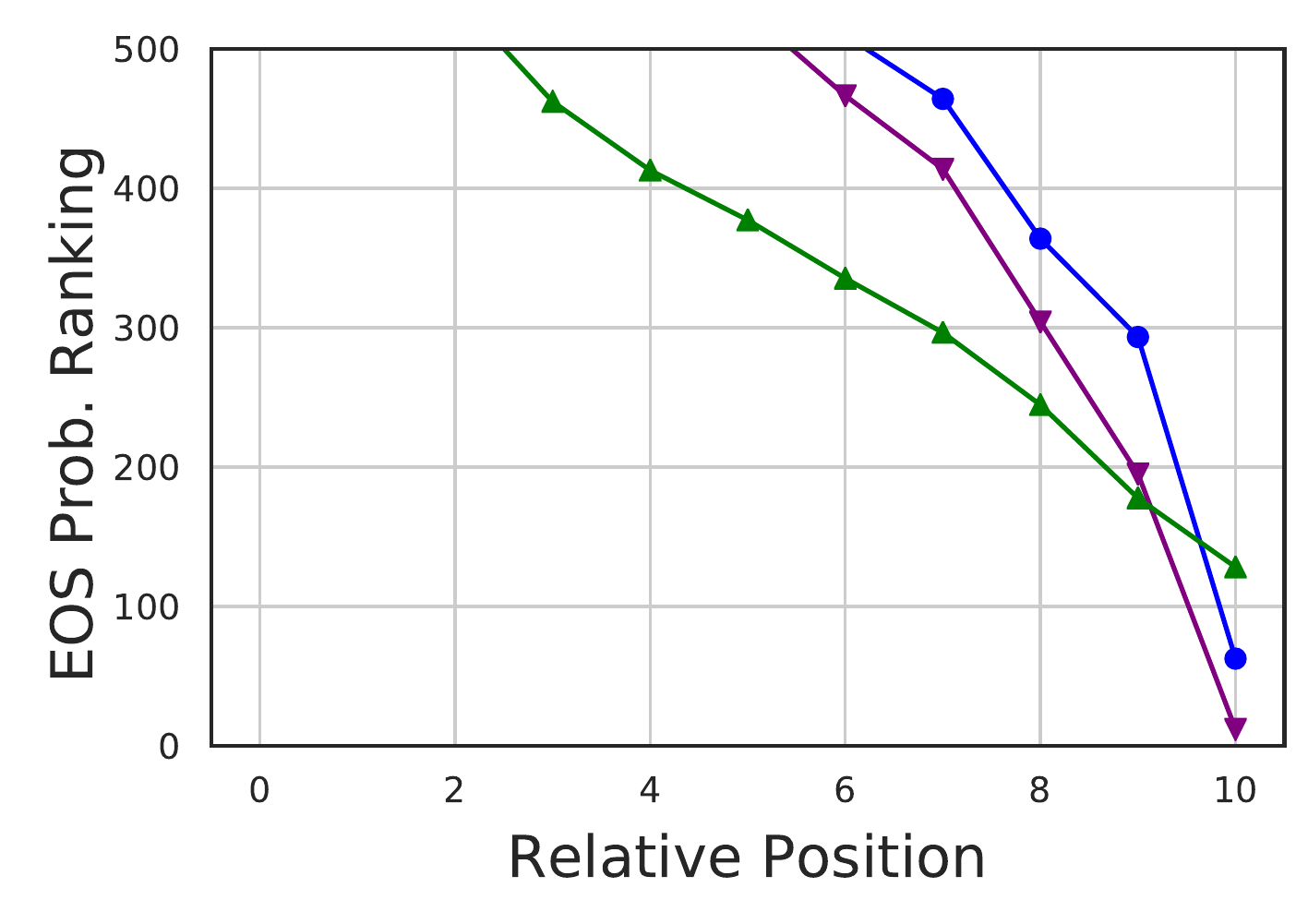}\label{fig:local}}
    \caption{Relationships between paragraph relative position and the ranking of the \eos~probability predicted by the last token of each paragraph.}
    \label{fig:appendix_eos}
    \vspace{-4mm}
  \end{figure*}

\begin{figure*}[ht!]
  \centering \small
    \subfigure[Topic relevance]
    {\begin{tabular}{lm{0.4cm}m{0.4cm}m{0.4cm}}
      &None &\sep~&\eop~\\
      None & &\cellcolor[HTML]{cccccc}43 &\cellcolor[HTML]{cccccc}45 \\
      \sep~&\cellcolor[HTML]{4285f4}\color{white}\textbf{57} & &\cellcolor[HTML]{cccccc}52 \\
      \eop~&\cellcolor[HTML]{4285f4}\color{white}\textbf{55} &\cellcolor[HTML]{cccccc}48 & \\
      \end{tabular}}
    \subfigure[Fluency]
    {\begin{tabular}{lm{0.4cm}m{0.4cm}m{0.4cm}}
      &None &\sep~&\eop~\\
      None & &\cellcolor[HTML]{cccccc}47 &\cellcolor[HTML]{cccccc}53 \\
      \sep~&\cellcolor[HTML]{cccccc}53 & &\cellcolor[HTML]{cccccc}54 \\
      \eop~&\cellcolor[HTML]{cccccc}47 &\cellcolor[HTML]{cccccc}46 & \\
      \end{tabular}}
    \subfigure[Ending quality]
    {\begin{tabular}{lm{0.4cm}m{0.4cm}m{0.4cm}}
      &None &\sep~&\eop~\\
      None & &\cellcolor[HTML]{cccccc}37 &\cellcolor[HTML]{cccccc}34 \\
      \sep~&\cellcolor[HTML]{4285f4}\color{white}\textbf{63} & &\cellcolor[HTML]{cccccc}49 \\
      \eop~&\cellcolor[HTML]{4285f4}\color{white}\textbf{66} &\cellcolor[HTML]{cccccc}51 & \\
      \end{tabular}}
    \subfigure[Overall preference]
    {\begin{tabular}{lm{0.4cm}m{0.4cm}m{0.4cm}}
      &None &\sep~&\eop\\
      None & &\cellcolor[HTML]{cccccc}42 &\cellcolor[HTML]{cccccc}47 \\
      \sep~&\cellcolor[HTML]{4285f4}\color{white}\textbf{58} & &\cellcolor[HTML]{cccccc}50 \\
      \eop~&\cellcolor[HTML]{cccccc}53 &\cellcolor[HTML]{cccccc}50& \\
      \end{tabular}}
    \caption{Average percentage of systems in row outperform system in column. Results are normalized without considering the Cannot Distinguish examples. }
    \label{fig:he_score}
  \end{figure*}
  
\begin{figure*}[ht!]
  \centering \small
    \subfigure[Topic relevance]
    {\begin{tabular}{lm{0.5cm}m{0.5cm}m{0.5cm}}
      &None &\sep~&\eop\\
      None & &\cellcolor[HTML]{cccccc}0.89 &\cellcolor[HTML]{cccccc}0.69 \\
      \sep~&\cellcolor[HTML]{cccccc}0.89 & &\cellcolor[HTML]{cccccc}0.70 \\
      \eop~&\cellcolor[HTML]{cccccc}0.69 &\cellcolor[HTML]{cccccc}0.70 & \\
      \end{tabular}}
    \subfigure[Fluency]
    {\begin{tabular}{lm{0.5cm}m{0.5cm}m{0.5cm}}
      &None &\sep~&\eop~\\
      None & &\cellcolor[HTML]{cccccc}0.61 &\cellcolor[HTML]{cccccc}0.60 \\
      \sep~&\cellcolor[HTML]{cccccc}0.61 & &\cellcolor[HTML]{cccccc}0.65 \\
      \eop~&\cellcolor[HTML]{cccccc}0.60 &\cellcolor[HTML]{cccccc}0.65 & \\
      \end{tabular}}
    \subfigure[Ending quality]
    {\begin{tabular}{lm{0.5cm}m{0.5cm}m{0.5cm}}
      &None &\sep~&\eop~\\
      None & &\cellcolor[HTML]{cccccc}0.68 &\cellcolor[HTML]{cccccc}0.63 \\
      \sep~&\cellcolor[HTML]{cccccc}0.68 & &\cellcolor[HTML]{cccccc}0.71 \\
      \eop~&\cellcolor[HTML]{cccccc}0.63 &\cellcolor[HTML]{cccccc}0.71 & \\
      \end{tabular}}
    \subfigure[Overall preference]
    {\begin{tabular}{lm{0.5cm}m{0.5cm}m{0.5cm}}
      &None &\sep~&\eop~\\
      None & &\cellcolor[HTML]{cccccc}0.52 &\cellcolor[HTML]{cccccc}0.34 \\
      \sep~&\cellcolor[HTML]{cccccc}0.52 & &\cellcolor[HTML]{cccccc}0.51 \\
      \eop~&\cellcolor[HTML]{cccccc}0.34 &\cellcolor[HTML]{cccccc}0.51& \\
      \end{tabular}}
    \caption{Fleiss' kappa $\kappa$~\cite{fleiss1971measuring} for the reliability of raters' agreement. The interpretation of $\kappa$'s value should be: poor agreement~($<0$), slight agreement~(0.01\textendash0.2), fair agreement~(0.21\textendash 0.4), moderate agreement~(0.41\textendash0.6), substantial agreement~(0.61\textendash0.8), and almost perfect agreement~(0.81\textendash1).}
    \label{fig:he_kappa}
  \end{figure*}

\section{Results}
\label{sec:results}
The results of different settings of utilizing paragraph information (ParaType) are shown in Table~\ref{table:results_essay} and Table~\ref{table:results_story}: concatenating all paragraphs into an uninterrupted sequence~(None); concatenating all paragraphs with ``\textbackslash n'' as the paragraph separator~(\sep~\nl); concatenating all paragraphs with a new token ``[\sep]'' as the paragraph separator~(\sep~DIY); appending ``\textbackslash n'' to the end of each paragraph~(\eop~\nl); appending a new token ``[\eop]'' to the end of each paragraph~(\eop~DIY).

\smallskip \noindent \textbf{Automatic Metrics.}
For Chinese essay generation, since Chinese-GPT2 is pre-trained without any special tokens(\eos/\eop/\sep), it will keep generating until meet the max length limitation without fine-tuning. 
In this case, we first compare None models fine-tuned with and without \eos~in Table \ref{table:results_essay}.
We can find that both T PPL~(-) and BLEU scores are better with \eos.
However, even fine-tuned with \eos, only 76.41\% generated texts can end with \eos.
After adding \eos, the \eos~PPL plunges from 22.15 to 2.74 and the \eos\% rising from 76.41 to 93.07, indicating that more generated essays end with the \eos~after learning to end paragraphs.
The BLEU scores are also improved.
It should be noted that the BLEU score is affected by the length of the text. 
We further truncate all generations with the length of the ground-truth story to calculate the truncated BLEU scores which are detailed in  Appendix~\ref{appendix: t-bleu} and the overall trending is consistent.
Finally, the ground-truth essays get 41.2 DIST1 and 82.65 DIST2, which means \eos~DIY achieves the closest DIST scores to the ground-truths.

On the other hand, English GPT2 is pre-trained with \eos~and line break \nl. 
Hence, we first compare GPT2 fine-tuned without \nl, with \nl, and with new token ``[\sep]''/``[\eop]''.
According to the Table \ref{table:results_story}, we can find that the fine-tuned GPT2 with \nl~as \sep~achieves the best results on word and token level perplexity metrics.
Compared with the model fine-tuned without paragraph information, all the models with \eop/\sep~achieve better BLEU scores. 
We further report the length truncated BLEU scores in Appendix~\ref{appendix: t-bleu}. The overall trending is consistent.
As for diversity score, the DIST1 and DIST2 of the ground-truth stories are 50.23 and 85.07, and the \sep~\nl~is the most close one.
In addition to the better PPL and BLEU score, we find that learning to end paragraphs can benefit the prediction of \eos.
The \eop~DIY achieves the lowest \eos~PPL and all models trained with \eop/\eos~achieve better \eos~PPL than model without paragraph information, except the \eop~\nl. 
This observation indicates that GPT2 tends not to generate the \eos~following the \nl~even after fine-tuning, but it can learn better \eos~with the help of a new \eop~token.

\begin{figure*}[h!]
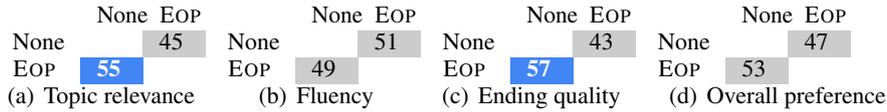

  \centering \small
    \subfigure[Topic relevance]
    {\begin{tabular}{lm{0.4cm}m{0.4cm}}
      &None &\eop~\\
      None & &\cellcolor[HTML]{cccccc}45 \\
      \eop~&\cellcolor[HTML]{4285f4}\color{white}\textbf{55} &  \\
      \end{tabular}}
    \subfigure[Fluency]
    {\begin{tabular}{lm{0.4cm}m{0.4cm}}
      &None &\eop~\\
      None & &\cellcolor[HTML]{cccccc}51 \\
      \eop~&\cellcolor[HTML]{cccccc}49 &  \\
      \end{tabular}}
    \subfigure[Ending quality]
    {\begin{tabular}{lm{0.4cm}m{0.4cm}}
      &None &\eop~\\
      None & &\cellcolor[HTML]{cccccc}43 \\
      \eop~&\cellcolor[HTML]{4285f4}\color{white}\textbf{57} &  \\
      \end{tabular}}
    \subfigure[Overall preference]
    {\begin{tabular}{lm{0.4cm}m{0.4cm}m{0.4cm}}
      &None &\eop~\\
      None & &\cellcolor[HTML]{cccccc}47 & \\
      \eop~&\cellcolor[HTML]{cccccc}53 & & \\
      \end{tabular}}
    \caption{Average percentage of systems in row outperform system in column.}
    \label{fig:he_score_en}
  \end{figure*}
  
\begin{figure*}[h!]
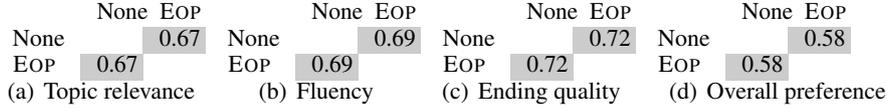

  \centering \small
    \subfigure[Topic relevance]
    {\begin{tabular}{lm{0.4cm}m{0.4cm}}
      &None &\eop~\\
      None & &\cellcolor[HTML]{cccccc}0.67 \\
      \eop~&\cellcolor[HTML]{cccccc}0.67 &  \\
      \end{tabular}}
    \subfigure[Fluency]
    {\begin{tabular}{lm{0.4cm}m{0.4cm}}
      &None &\eop~\\
      None & &\cellcolor[HTML]{cccccc}0.69 \\
      \eop~&\cellcolor[HTML]{cccccc}0.69 &  \\
      \end{tabular}}
    \subfigure[Ending quality]
    {\begin{tabular}{lm{0.4cm}m{0.4cm}}
      &None &\eop~\\
      None & &\cellcolor[HTML]{cccccc}0.72 \\
      \eop~&\cellcolor[HTML]{cccccc}0.72 &  \\
      \end{tabular}}
    \subfigure[Overall preference]
    {\begin{tabular}{lm{0.4cm}m{0.4cm}m{0.4cm}}
      &None &\eop~\\
      None & &\cellcolor[HTML]{cccccc}0.58 & \\
      \eop~&\cellcolor[HTML]{cccccc}0.58 &  & \\
      \end{tabular}}
    \caption{Fleiss' kappa $\kappa$~(ranging from -1.0 to 1.0) for the reliability of raters' agreement. }
    \label{fig:he_kappa_en}
  \end{figure*}
We further compared the relations between \eos~and different \eop/\sep, which is shown in Figure \ref{fig:appendix_eos}. 
The horizontal axis represents the relative paragraph index, 0 means the beginning paragraph and 100 means the last paragraph of the story. 
The vertical axis represents the ranking position of the \eos~probability among all tokens in the vocabulary predicted by the last token of each paragraph. 
As \eos~should only be predicted by the last token of the last paragraph, the ranking at 100 should be higher and the other position should be lower. 
According to Figure \ref{fig:global}, all models rank \eos~higher as the paragraph index increasing. 
\eop~works better than \sep~as the \eop~models rank \eos~higher at the 100th position and lower on the other positions, which can be seen from Figure \ref{fig:local}.

\smallskip \noindent \textbf{Human Evaluations.}
Human evaluation results are shown in Figure~\ref{fig:he_score}. 
Each cell represents the percentage of the examples that the row system wins the column system on. 
Cells will be filled in blue if the row system outperforms the column system over 10\%.
It should be noted that Cannot Distinguish examples are skipped when counting winners for these figures.
From Figure~\ref{fig:he_score}, we can first find that learning to end paragraphs leads to better ending quality: 63\% and 66\% results are rated better when comparing \sep/\eop~with None systems, while only 37\% and 34\% results are rated better for None system. 
We also find that \eop's text endings are slightly better than \sep.
This is consistent with \eos~PPL and \eos\% results in Table~\ref{table:results_essay}.
Although \sep~wins on topic relevance and fluency, the overall preference of \sep~compared with \eop~is 50\%, which means these two systems are similar for human rates.
Besides, we also find that \sep~and \eop~achieve better topic relevance and overall preference.
For fluency, there is no significant difference among different systems.

We also report Fleiss' kappa $\kappa$ in Figure~\ref{fig:he_kappa}, to access the reliabilities of raters' agreement. 
$\kappa<0$ means poor agreement, and $\kappa\sim(0.6,0.8)$ means substantial agreement.
From this figure, we can find that most of them fall into the substantial agreement group.
The overall preference falls into moderate agreement, because this metric is more subjective than the others. 

Human evaluation results for WritingPrompts are shown in Figure~\ref{fig:he_score_en} and Figure~\ref{fig:he_kappa_en}.
Assessors still prefer model fine-tuned with \eop~rather than without \eop/\sep.
  
\smallskip \noindent \textbf{Case Study.}
We further conduct case study and detailed in Appendix~\ref{case study}.
The most important observation is that, without \eop, the beginning of the generation is more relevant to the end of the input prompt, but the more it generates, the poor quality is. 
While the generator with \eop~can generate multiple paragraphs related to the input with a reasonable ending but each paragraph is more independent than human writings.

%% file: sections/conclusion.tex
\section{Conclusion}
In this paper, we have demonstrated that \eop~and \eos~information helps generating better text.
Chinese GPT2 and English GPT2 are two existing models pre-trained without and with \eop~respectively, which provides a perfect platform for our proposed experiments.
On the ChineseEssay dataset, the text generation when fine-tuned with \eop~and \eos~information is significantly improved. 
On the other hand for the English task, although (English) GPT-2 was trained with \nl~which serves as \eop~to some degree, learning to end paragraphs can still benefit the story generation in terms of automatic metrics and human evaluation results.



%% file: sections/appendix.tex
\clearpage
\appendix
\section*{Appendix Overview}
In this supplementary material, we provide additional experimental results of truncated BLEU score in Appendix \ref{appendix: t-bleu}, and several generations in Appendix \ref{case study}.
\input{appendix/truncated_bleu.tex}

\input{appendix/case_study.tex}

%% file: appendix/truncated_bleu.tex
\begin{table*}[bp]
  \centering \small
  \begin{tabular}{l|c|c|c|c|c}
  \hline
\textbf{ParaType}&\textbf{FT}&\textbf{\eos}&\textbf{T-BLEU1}&\textbf{T-BLEU2}&\textbf{Avg.Length}\\
  \hline
  \multirow{3}{*}{None} & No &No &42.6&9.0&814.1\\ 
  \cline{2-6}
   & Yes&No &48.0&11.6&793.5\\ 
   \cline{2-6}
   & Yes&Yes &49.2&11.9&632.0\\ 
  \hline
  \sep~DIY& Yes&Yes&49.4&12.0&576.7\\ 
  \hline
  \eop~DIY& Yes&Yes&49.6&12.0&535.4\\ 
  \hline
  GroundTruth& -&- &-&-&571.3 \\
  \hline
  \end{tabular}
  \caption{\label{table:results_tb_essay}Average length and truncated BLEU scores of different models on ChineseEssay dataset.} 
  \vspace{-2mm}
  \end{table*}
  
\begin{table*}[bp]
  \centering \small
  \begin{tabular}{l|c|c|c|c}
  \hline
  \textbf{ParaType}&\textbf{FT}&\textbf{T-BLEU1}&\textbf{T-BLEU2}&\textbf{Avg.Length}\\
    \hline
    None & Yes &36.6&5.3&392.7\\ 
    \hline
    \sep~\nl~& Yes &37.6&5.0&367.4\\ 
    \hline
    \eop~\nl~& Yes &37.3&5.0&368.0\\
    \hline
    \sep~DIY & Yes &38.6&8.0&385.7\\ 
    \hline
    \eop~DIY& Yes &38.6&8.1&379.6 \\
    \hline
    GroundTruth& - &-&-&369.3 \\
  \hline
  \end{tabular}
  \caption{\label{table:results_tb_story}Average length and truncated BLEU scores of different models with finetuning on WritingPrompts dataset. } 
  \end{table*}

\section{Truncated BLEU Score}
\label{appendix: t-bleu}
The BLEU score is easily affected by the length of text, where a short text might achieve a higher BLEU score than a long text. 
The average lengths of the texts generated from different methods are shown in Table~\ref{table:results_tb_story} and Table~\ref{table:results_tb_essay}.
An intuitive metric for this problem is the truncated BLEU~(T-BLEU) score.

To get the T-BLEU score, we first truncate the generated text with the length of its corresponding ground-truth text.
Then, the BLEU score of the truncated text is the T-BLEU score.

As we can see from  Table~\ref{table:results_tb_story} and Table~\ref{table:results_tb_essay}, although the BLEU score improvements of \eop/\sep~become less significant on the Chinese dataset, the overall trending is similar with the normal BLEU scores.

%% file: appendix/case_study.tex
\section{Case Study}
\label{case study}
We first conduct case studies with Chinese GPT2. 
Case \ref{case:zh-1} and Case \ref{case:zh-2} are two cherry-picked examples.

The prompt of the first example Case \ref{case:zh-1} is about the author's teacher. 
After finetuning without paragraph information, we can see that the generated continuation is related to the given prompt but pays too much attention to the gifts instead of the teacher, and generating something about the financial problem in the beginning. 
Although the middle portion of the continuation is well written, the latter half part is poor, incomplete and hard to be understood. 
In contrast, the continuation generated with \eop~is much better, although with minor errors of word choice. 
Besides, the ending of the latter one is much better as the former one just keeps generating until reaches the maximum length. 

Similar trending can be found in the second example Case \ref{case:zh-2}. 
According to our observation, without \eop, the beginning of the generation is more relevant to the end of the input prompt but the more it generates, the poor the quality is. 
While the generator with \eop~can generate multiple paragraphs related to the input with a reasonable ending but each paragraph is more independent than human writings.

We further investigate the story generation results. 
Two different cases are shown in Case \ref{case:en-1} and Case \ref{case:en-2}. 
In the first case, the generation of DIY \eop~is better than \nl~\sep, as it mentions the aliens multiple times and also generates the time ``100 years'' while the other one only talks about the planet. 
In Case \ref{case:en-2}, \nl~\sep~is better than DIY \eop. 
The generator fine-tuned without paragraph information is the poorest in these two cases, which is consistent with our experimental results and indicates the paragraph information can benefit the story generation.

\clearpage
\onecolumn
\subsection{Case 1 of Chinese-GPT2}
\label{case:zh-1}
\begin{figure}[!htb]
\centering \small
\fbox{
\begin{minipage}{.95\textwidth}
\textbf{Input prompt:}\\
\begin{CJK*}{UTF8}{gbsn} 
老师，就像辛勤的园丁，每天都向我们灌溉知识。
老师，就像一支红烛，照亮了我们，燃尽了自己。
老师，就像那苍天大树，而我们，就是树下的一棵小草。
老师，就像那蓝天上的白云，而我们，就是自由翱翔的小鸟。
老师，就像那无尽的海水，而我们，就是哪快乐畅游的小鱼。
老师，就像那广袤无垠的大草原，而我们，就是哪草原上奔腾的骏马。
\\
\\
每年快到教师节的时候，我都会无比的感谢老师，因为，有了老师，我们就是幸福的人，比那些在大山上不了学的，我们是多么的幸运。
有了老师，我们就是有了尾巴的鱼，在无尽的知识里畅游。
有了老师，我们就像有了翅膀的雄鹰，在蔚蓝的天空中，自由翱翔。
\\
\\
我今年小学毕业，我离开了上了6年的小学，那里有熟知的人，熟悉的教室，可是那就是过去，是我回忆的地方，每年的教师节，我听到我们班主任说的最多的一句话就是：下次别再送了礼物了，我要的不是那些礼物，我要的是你们学习的态度，如果，你们要送的话，就送些手工做的，可以旧物利用，比如，叠个千纸鹤了，叠个星星了，都可以，不要用钱买，有那份钱，也可以买些学习用品。
\end{CJK*}  \\
\textbf{Brief Translation:}\\
I am grateful to my teacher. The teacher's guidance to us is essential like the tail to the fish and the wings to the eagle. This year, I have graduated from primary school which is a place full of good memories. I still remember that every Teachers' Day, my teacher told us not to waste money to prepare gifts for her, as learning together is the best gift. 
 \end{minipage} }
   \caption{Chinese input prompt and English translation.} 
\end{figure}

\begin{figure}[!htb]
\centering \small
\fbox{
\begin{minipage}{.95\textwidth}
\textbf{Output of Chinese-GPT2 fine-tuned without \eop:} \\
\begin{CJK*}{UTF8}{gbsn}
我记得有一次，班主任说有同学家里穷，需要资助，因为我们很多\textbf{同学都没有孩子}({unreasonable})，一般都是自己挣钱，所以，我们同学的爸爸妈妈都说："你们没钱没权，家里的\textbf{爸妈都没有孩子}({unreasonable})，这对我们来说是一种压力，因为我们的父母的学历都比你们高，就算我们再穷，我们也要好好读书，不然到时候就没的读了。"我听到这句话，心里像扎了根针一样，痛。\textbf{我希望老师知道我的心声，并对我进行语文课程的辅导，让我能更加好的理解课文。谢谢老师，让我在欢笑中学习知识，学会感恩，学会感谢辛苦付出的老师。老师，我们是彩虹，落在心里最美的地方，遇到，便是老天最大的幸运，让我们用最美的彩虹，每天绽放最灿烂的彩虹，为我们的校园增添一缕鲜花} (\textit{good continuation but should stop here.})  奪す她，虽已不在人世，但我永远不会忘记她，为了我，我愿用自己全部的青春献出自己的真心，把我对老师的感谢埋进心底，告诉父母：我对老师的感恩，与幸福一样宝贵。笨呀，笨死了，我送了一本漫画书给她。我看到了她泪眼婆娑，一脸的疲惫，并且嘴里还喃喃着："对不起，我是单亲家庭，不能送你一本书了。"我哽咽着，扶了扶眼镜，他一把拽住了我的衣服："这有什么，我都已经这样了，还想着你，你居然还想走....."我再三道谢，沉默地笑了。笨呀，有什么好抱怨的，我不知道以后怎么办，以后，我愿不(\textit{incomplete})
\end{CJK*} \\
\textbf{Modified Google Translation:} \\
I remember a time when my teacher told us that one of our classmates need financial assistance, because many of \textbf{our classmates had no children}(\textit{unreasonable}) and generally made their own money, so our classmates' moms and dads said," You have no money and no rights. \textbf{Parents has no children}(\textit{unreasonable}) , which is a stress for us, because our parents have higher academic qualifications than you. Even if we are poor, we must study well, otherwise we will have no chance of studying. " At this point, my heart hurts like be inserted in a needle.\textbf{ I hope that the teacher will know my minds and provide me with Chinese language tutoring so that I can better understand the text. Thank you teacher, let me learn knowledge happily, learn to be grateful, learn to thank the hard-working teacher. Teacher, we are the rainbow. We fall in the most beautiful place in our hearts. When we meet, it is the greatest luck. Let us use the most beautiful rainbowe brightest rainbow every day and add a ray of flowers to our campus. }(\textit{good continuation but should stop here.}) [garbled tokens], she has passed away but I will never forget her. I told my mother that I am grateful to my teacher. I don't know what to do. In the future, I hope not(\textit{incomplete})\\
 \end{minipage} }
   \caption{Generation from Chinese-GPT2 fine-tuned without \eop.} 
\end{figure}
\clearpage
\begin{figure}[!htb]
\centering \small
\fbox{
\begin{minipage}{.95\textwidth}
\smallskip \noindent \textbf{Output of Chinese-GPT2 fine-tuned with \eop:}\\
\begin{CJK*}{UTF8}{gbsn}
老师，这是一个实实在在的老师，她知识渊博，课堂上，她激发我们的斗志，课下，她严格，刻苦，为我们全班同学排忧解难，可是在我的生活中，我最敬佩的老师是一位母亲，老师，是用爱心来温暖，去回报，用知识来创造美丽幸福的人。有了老师的教导，我们就像上帝创造出来的天使，是最纯净的天空，没有感伤，没有痛苦，每一天，我都在老师的呵护下，享受\textbf{天伦之乐}(\textit{unreasonable})，是您指引着我们，走向未来，是您为我们付出了许多，是您使我们在知识的海洋里健康地成长。\\
\\
曾经有一位心灵手巧的老师，让我们学会手工活，学会穿衣服，学会修剪指甲，收拾桌椅，学会穿衣服，学会洗衣服......她就是我们的语文老师。说真的，我心里很感激我的语文老师。她教会我们词语的讲解和组合，教会我们汉字的发音，曾经我的心情很低落，她的安慰，为我指点迷津，让我重新燃起了奋斗的斗志。\\
\\
我会好好的学习，才不会辜负老师的期望呢！[\eos]
\end{CJK*}\\
\textbf{Modified Google Translation:}\\
Teacher, this is a trustworthy teacher. She is very knowledgeable. In class, she inspires our fighting spirit. After class, she is strict and diligent. She solves problems for our classmates. The teacher is a mother, a teacher, a person who warms with love, returns, and uses knowledge to create beautiful and happy people. With the teaching of the teacher, we are like the angel created by God. It is the purest sky, no sentimentality, no pain. Every day, I am under the care of the teacher and enjoy the \textbf{family fun}(\textit{unreasonable}). It is you who lead us to the future, it is you who paid a lot for us, and it is you who made us grow up healthy in the ocean of knowledge.

There used to be a handy teacher who taught us how to work by hand, how to dress, how to trim nails, how to clean tables and chairs, how to dress, how to wash clothes ... she is our Chinese language teacher. Seriously, I am very grateful to my Chinese teacher. She taught us the explanation and combination of words, taught us the pronunciation of Chinese characters. And when I was depressed, her consolation gave me pointers to help me rekindle my fighting spirit.

I will study hard so that I will not disappoint the teacher's expectations!
 \end{minipage} }
   \caption{Generation from Chinese-GPT2 fine-tuned with \eop.} 
\end{figure}

\clearpage
\subsection{Case 2 of Chinese-GPT2}
\label{case:zh-2}
\begin{figure}[!htb]
\centering \small
\fbox{
\begin{minipage}{.95\textwidth}
\smallskip \noindent \textbf{Input Prompt:}\\
\begin{CJK*}{UTF8}{gbsn} 
那时，正值夏季，烈日炎炎。\\
\\
我和我的朋友们正走回学校的路上。
\end{CJK*} \\
\textbf{Brief Translation:}\\
It was a hot summer.
I was on the way to walking back to school with my friends.
 \end{minipage} }
   \caption{Chinese input prompt and English translation.} 
\end{figure}

\begin{figure}[!htb]
\centering \small
\fbox{
\begin{minipage}{.95\textwidth}
\smallskip \noindent \textbf{Output of Chinese-GPT2 fine-tuned without \eop}\\
\begin{CJK*}{UTF8}{gbsn}
  \indent凉爽的风吹来，我的心也舒畅了些许。走出校门，看到一大群人在热火朝天地玩着什么，我不禁打了个冷战。只见有几个小孩子挎着摇摇篮，走在空地上，摇摇篮里塞满了不听话的人，吵吵嚷嚷地叫着，吵闹的人群一个接着一个地的在空地上来回飞舞。只见他们不管是乱扔垃圾的，还是在家里乱扔东西的，总有那么一两个人，让人看了都觉得讨厌，那几个男生，在那儿使出浑身解数，往那儿扔垃圾，有的还不止一次，他们真的是抓破了脑袋才钻到垃圾堆里啊！"听说那些事儿很早就埋在那儿了。"他们越说越起劲，像每个夏季都会发生的那样，还拿起了"帚，拿起了沙包。回过神来，脸已经晒得通红，汗珠直往外流。我呆呆地站在那儿，思绪飞到了几年前，那时的我才十三岁，现在的我才二十四岁！打来打去，什么也没有留下。唉，打得还挺勤快。我转过身，在树底下端详着他们，又想起了他们，他们可真是坏事当前，坏事晚来了许多啊！我继续打打我那张臭脸。我的心猛地一抽，丢下书包，飞奔出去。在那片空地上，零零星星的人有的在玩儿，有的正在玩呢！他们又跳又叫，不分是非地叫着。我突然看见一个不起眼的小孩儿，他满脸通红，眼睛红得像个灯泡，让人看了如沐春风。他正用眼睛注视着那只有倒在地的人，他下意识地往上看，一会儿，上面散落着一些塑料瓶子。他不知道，还在玩儿呢！我忽然有些感动，急忙跑过去，跟在他后面，问："他怎么了？为什么那么残忍。"小伙儿只是若无其事地回答："我觉得他太可怜了。"他倒也会说话，但是有些话让我怎么说也说不出来。他一双大大的眼睛仿佛在说："我这是为了家里，也是为了我们共同的孩子。现在，我是想退出，你们也别急，知道吗？要是真和这些小孩闹僵了，家里的学习教材再翻出来，你们就完了！"看着小伙儿那可怜的小眼睛，心里一阵酸楚，我不禁流下了眼泪。每次吵完架后，他总爱说："我不怪你，可你就这样放弃我吧！我没钱啊！"唉，啊！唉！哎！气息都快从嗓子里出来了啊！唉！唉！唉！唉！那时，我仿佛听见了他的声音："哎，孩子，爸爸的房屋可能拆迁了，你就住在那吧。等拆迁后我送你吧！我会好好学习的！"那声音大得连我自己都听不见，我的心仿佛像灌了铅似的。我也哭了，但没办法，谁叫人生是漫长的呢？那时我只会低着头走路，却找不到一根自来水。我生着闷气，仿佛快要窒息了。我急得满头大汗，冲到楼下，不停地喊："救救我！救救我！"唉，我都快把脸憋得通红。这时，对面来了一辆出租车，我毫不犹豫地将
\end{CJK*} \\
\textbf{Modified Google Translation:}\\
The cool wind blew, and my heart was relieved a little. I walked out of the school gate and saw what a large group of people were doing something. I couldn't help fighting a cold war. I saw a few children carrying the cradle and walking on the open space. The cradle was full of disobedient people, screaming loudly, and the noisy crowd flew back and forth on the open space one by one. I saw that whether they were littering or throwing things at home, there were always one or two people that made people feel annoying when they saw it. Those boys, they tried to do everything and throw garbage there. Some more than once, they really broke their heads and got into the garbage dump! "I heard that those things were buried there very early." They said more and more vigorously, as they would happen every summer, and took up "Broom, took up the sandbag. When I looked back, my face was already sunburned It was red, sweat was flowing out. I stood there blankly, my thoughts flew a few years ago, when I was only 13 years old, and now I am only 24 years old! Stay. Alas, I played very diligently. I turned around, looked at them under the tree, and thought of them again. They are really bad things, and bad things come a lot late! I continue to beat my bad face . My heart twitched, dropped my schoolbag, and ran out. On the open space, some people were playing and some were playing! They jumped and called, regardless of right or wrong. Screaming. I suddenly saw an inconspicuous child, his face flushed, his eyes red like a light bulb, making people look like a spring breeze. He was staring at the only person who fell to the ground with his eyes, he subconsciously went up Look, for a while, there are some plastic bottles scattered on it. He doesn't know, he's still playing! I suddenly Some moved, quickly ran behind him, asked: "how he was? Why so cruel. "The boy just answered casually:" I think he's too pathetic. "He would speak, but some words made me unable to say what he said. His big eyes seemed to be saying," This is for my family and for our common children. Now, I want to quit. Don't worry, you know? If you are really upset with these children, you will find out the learning materials at home, and you are finished! "Looking at the poor little eyes of the guy, I felt sore in my heart, and I couldn't help crying. After each quarrel, he always said," I don't blame you, but you just give up on me! I have no money! "Oh, ah! Uh! Hey! The breath is almost coming out of my throat! Uh! Uh! Uh! Uh! Then I seemed to hear his voice:" Oh boy, my father's house may be demolished, You live there. I'll send you after the demolition! I will study hard! "That voice was so loud that I couldn't even hear it myself, my heart seemed to be filled with lead. I also cried, but there was no way, who made life long? At that time, I just walked with my head down, But I couldn't find a tap. I was sulking, as if I was about to suffocate. I was so sweaty that I rushed downstairs and kept shouting, "Save me! help me! "Well, I'm almost flushing my face. At this moment, a taxi came across, and I didn't hesitate to\\

 \end{minipage} }
   \caption{Generation from Chinese-GPT2 fine-tuned without \eop.} 
\end{figure}
\clearpage
\begin{figure}[!htp]
\centering \small
\fbox{
\begin{minipage}{.95\textwidth}
\smallskip \noindent \textbf{Output of Chinese-GPT2 fine-tuned with \eop}\\
\begin{CJK*}{UTF8}{gbsn}
  \indent我们都还在上课，突然一阵急促的车铃声传来，我们低下了头，拿出粉笔，认真地练习。

  \indent窗外的大树上，有一个乞丐，他向路人求助，却得不到人帮助，放弃了乞丐的工作，独自一人走了。他的脸上，永远挂着笑，脸颊上，微微露出皱纹。

  \indent火车缓缓开过，熙熙攘攘的人群和我们汇成一片。那时，我正在大口大口地吃着零食。突然，看到一个乞丐，他颤巍巍地走了过来，我急忙拦住他："小心，别被车轧到。"他却语气坚决，关切地问："您怎么这么着急，知不知道车辆出轨了？我们的工作又不是那么危险，怎么可以冒险出来帮助别人呢？您一天到晚忙着赚钱，却不关心自己的健康，给钱这么不讲道理，真可怜。""啊，这是什么人啊？拿什么关心呀？这是个老人啊！他这样只顾自己，在哪打点零工，卖什么不关心。"一旁的马路上，几个小伙子正在大口大口地讨论着。我不禁流下了眼泪。含着眼泪，向其中一个乞丐致谢，这个老人见了我们，又弯下腰，打开自行车的后座，把钱给了我们，又对我们说："不要再来烦我们了，我们的工作就是很危险的，也很需要钱的，不如来为国家做点力所能及的事吧。"我的泪水模糊了我的眼睛。

  \indent慢慢地，车流渐渐远去，但有些在星星点点的人群里，依然在四处寻找着那个乞丐。夜深人静了，我一人在街上漫步，深深地感受到社会的冷漠和人性的丑恶。我的心又忍不住泛起阵阵潮湿的雨珠。一件一件事情，令我感到我自己的不幸。随后，雨又愈发大了。

  \indent我环顾四周，在最好的车站等上了车，我并没有去接那乞丐。直到车子终于来了，我才有些不耐烦了。不是不想让别人帮我，而是那些恶作剧都是不成熟的表现。很多时候，对人也是冷漠的，甚至是冷嘲热讽。我有些生气，却又难以控制着。游戏在幸福的雨幕里。以后每当看见乞丐，我总是感到心酸。

  \indent在接下来的日子里，我不再光顾街头，但行人仍多。我依然不理会任何人的言语，回家却无法挽回，那个老人又怎么会有人关心呢？

  \indent时间会冲淡一切，但我不想放弃，就算老天在我身边，我还是会努力，继续做好自己，虽然结果不尽如人意，但我依旧要用自己的行动报答社会。[\eos]
\end{CJK*}\\
\textbf{Modified Google Translation:}\\
We were all still in class, and suddenly there was a rush of car bells. We lowered our heads, took out chalk, and practiced seriously.

In the big tree outside the window, there was a beggar who asked for help from passersby, but could not get help. He gave up the work of the beggar and left alone. There was always a smile on his face, and a slight wrinkle appeared on his cheek.

The train passed slowly, and the bustling crowd merged with us. At that time, I was eating snacks with a big mouthful. Suddenly, when I saw a beggar, he came tremblingly, and I hurriedly stopped him: "Be careful, don't get caught in the car." But he resolutely asked, "Why are you so anxious, do you know the car is out of track Now? Our work is not so dangerous, how can we take risks to help others? You are busy making money all day, but you don't care about your health. It 's so pitiful to give money so unreasonably. "" Ah, what is this? Man? What do you care about? This is an old man! He only cares about himself like this, where to do odd jobs and don't care what he sells. "On the side of the road, a few young men were talking in a big mouth. I can't help crying. With tears, thanks to one of the beggars. The old man saw us, then bent down, opened the back seat of the bicycle, gave us money, and said to us, "Don't bother us anymore, our work is very dangerous. I also need money, so let's do something for my country. "My tears have blurred my eyes.

Slowly, the traffic gradually went away, but some of the little crowds were still looking for the beggar. The night was quiet, and I was walking on the street alone, deeply feeling the indifference of society and the ugliness of human nature. My heart could not help but burst with wet raindrops. One thing at a time makes me feel my own misfortune. Afterwards, the rain became heavy again.

I looked around and waited at the best station. I didn't pick up the beggar. I didn't get impatient until the car finally came. It's not that you don't want others to help me, but that those pranks are immature. Many times, it is indifferent to people, even ridiculous. I'm a little angry, but hard to control. The game is in the rain of happiness. Whenever I see a beggar, I always feel sad.

In the following days, I stopped patronizing the streets, but there were still many pedestrians. I still ignore the words of anyone, but I ca n't restore it when I go home. How could anyone care about that old man?

Time will dilute everything, but I don't want to give up. Even if God is by my side, I will still work hard and continue to be myself. Although the results are not satisfactory, I still have to repay the society with my own actions.
 \end{minipage} }
   \caption{Generation from Chinese-GPT2 fine-tuned with \eop.} 
\end{figure}

\clearpage
\subsection{Case 1 of GPT2-117}
 \label{case:en-1}
\begin{figure}[!htb]
\centering \small
\fbox{
\begin{minipage}{.95\textwidth}
\textbf{Input prompt:}\\
Aliens come to Earth in hoping to wipe us out and take over the planet, but are honorable and gives us a fighting chance.  They announce they will return in 100 years so as to give Earth a chance to prepare and defend itself.  Neither side expected what the other would bring to war 
 \end{minipage} }
   \caption{English input prompt.} 
\end{figure}

\begin{figure}[!htb]
\centering \small
\fbox{
\begin{minipage}{.95\textwidth}
It was dark as the day began. Pure white, forever not knowing the night sky. This was the first time since their arrival that \textbf{ships} had been so dark, it was hard not to feel the tremors of terror as the drops of rain started falling. The rain started a minute after the \textbf{humans} had arrived. It started getting brighter as they piled up and felt the tremors of rain take their place. Another minute or two they saw them in the distance. One little shell, below them floated a little bigger. Behind that shell came another of them falling, one in each direction. Then, faster and faster the smaller the shell went. It was as if they were floating to each other, different colors and shapes sailing past each other in any direction, making the scene less momentary. The rain started to fall. The tall suits of \textbf{soldiers} patrolled the area. They didn't see what was happening but one was seeing some sort of change. He was slowly meeting the upper edge of the shell, directly under the soft snow. The rain started to fall and he was choking back the hail and it suddenly stopped flowing. He looked up at the Sun and the grass and saw the cloudless sky in front of him. He knew that had given him a new hope. He wasn't lying when he said that the atmosphere was starting to roll away, and just as he was about to turn around he felt a knock on his door. "I'm having a look at this. It's new... " " We'll wait. " " You have to go, have a look. " " Come with us." With that said the strange man took off his coat and sat on the porch of his house, waiting for the rain.
 \end{minipage} }
   \caption{Generation from GPT-117 fine-tuned without \eop/\sep.} 
\end{figure}

\begin{figure}[!htb]
\centering \small
\fbox{
\begin{minipage}{.95\textwidth}
A \textbf{planet} - there is no other more appealing than a\textbf{ peaceful planet}, \\
taking in the vast plains of green and dry, \\
allowing the native \textbf{species} to develop. \\
Among many other things, the greatest threat here was our neighbours as well, \\
hosting none but a single\textbf{ blue planet}. \\
With every news we received, while we was still, the green grass kept \\
tilling even the most \textbf{exotic plants}. \\
Against the odds, we agreed. This blue planet \\
is sure to die if it too falls, \\
as if the planet we are about to take hold of is not giving it. \\
Whether it is the region of lush land, \\
living in the shadow of endless sea, or the ones that lie beyond, \\
this \textbf{planet will feed our population} just as well. 
 \end{minipage} }
   \caption{Generation from GPT-117 fine-tuned with \nl~as \sep.} 
\end{figure}
\clearpage
\begin{figure}[!htb]
\centering \small
\fbox{
\begin{minipage}{.95\textwidth}
Robert stood in the room quietly. The pager in the edge of his TV blinked off as he contemplated what he was seeing. It was not his first visit by \textbf{alien ships}. Most of them had flown over the \textbf{galaxy in 882} since the foursome of an ancient race had first started transit in the Kardashev system. But they did not have a \textbf{star system} to conquer or destroy. \\
 The strain of building a bridge was far from comfortable as it had become. Even hands and feet were shaking as they dutifully lugged the countless cargo back to their vessel. \textbf{Each alien lacked a ship just like him}. Their ships would not be ready \textbf{for another 100 years}. Without time, they would be easily overwhelmed. \\
 The first of them came on us. The black, streaming in through a far wall. Its shape changing from green to orange as it accelerated. It came in waves across the curved surface. Its \textbf{metallic surface simply glowed off}. Robert had seen \textbf{the mouth of the creature. It had a back in it, almost reptilian. A spider-like body with a fangs that ripped through the flesh like a spider.} The mouth was growling with hunger in response. It roared as its muscles launched away from Robert to mate with him. \\
 The next twenty-three seconds went by with little delay. Robert led the old stranger down a hallway to his repair bay. He was quick to greet the human with baited breath and followed close behind him. He took the sidearm from underneath his suit and held it in front of his eyes as the humans tore at the metal. A metal band formed around the metal covered mouth. The human was tongue hanging off his neck. \\
 Without stopping, he freed his hands from the talons hanging around his head as he pressed the new badge on his hip. \\
 Robert opened his mouth to scream but his goal was to prevent the humans from catching on to what was happening. Robert began unwrapping a vial of orange liquid. He approached the canine and began licking it, waiting. The canine howled as Robert removed his mask and walked away. He had fifty percent of the tanker harvested and stored at a distance of about five million miles. \\
 The alien warship stopped in mid-flight and seemed to accelerate down again as Robert completed its tasks. He called back, letting out a whistle that was barely audible over the din of battle. Within a second the ship touched down at a small cove. The human ran over to him, mewling. \\
 "Not again. " Robert said. \\
 " Exactly, Mr. Tammoy. We have evidence of life support in here. " \textbf{The alien asked}. \\
 " And what. Was. All. You. Deposed. " Robert responded. \\
 " Quite literally. "\textbf{ The humanoid replied}. \\
 " So, just one moment. Can't you just give it a chance? " \\
 " Neither can anybody else." The creature continued, seemingly oblivious to its words and their implications. 
 Robert slowly brought the cockpit of the ship's other 'Crew' to the nub. He unlocked the pods and stepped into the airlock. The robot was gone. Robert sat in the cockpit, the cylindrical container behind him. He took a deep breath and looked out into the blackness, following the exact pattern of the vessel. The black flesh was always fascinating. More often than not, he didn't really understand what it meant. \\
 Robert walked past the humans towards his repair bay. It had not done much damage. He walked back past the humans and got into his pickup truck. With the last line of effort he pushed his way through the trio. 
 \end{minipage} }
   \caption{Generation from GPT-117 fine-tuned with DIY \eop.} 
\end{figure}

\clearpage
\subsection{Case 2 of GPT2-117}
   \label{case:en-2}
\begin{figure}[!htb]
\centering \small
\fbox{
\begin{minipage}{.95\textwidth}
\smallskip \noindent \textbf{Input Prompt:}\\
You were driving to work one day and your car suddenly started flying.
 \end{minipage} }
   \caption{English input prompt.} 
\end{figure}

\begin{figure}[!htb]
\centering \small
\fbox{
\begin{minipage}{.95\textwidth}
( got more into poetry / fiction themes/ before calling it Work ) My grandfather had 3 boxes of cake and a white sugar please Jane said "today is April 25th, and your seventh or eighth " I was in love by the looks of things, I hated her " Your father loves you at least nine times " I said " Oh, I know, it's really funny. " She looked at me with a confused expression " why would she do that " Then I began to back out of it like I told her I would. At that point Jane said something " no, she had a birthday today " and since I was the father of her child I felt really blessed for making her mother happy. We closed the front door and walked back into her house to eat cookies. You see, she almost did so fast, I was so pumped and pretty much took the bag to the kitchen for milk and honey, but here I was going to take everything home and spend it running errands. My wife moved back in a minute later, but I saw her and told her I was having this late so I told her " Yeah " " tell her I'll grab one at the bus stop tomorrow " She said you'll come back to bed early and eat one " goodbye " but I did hurry. My wife usually leaves early so \textbf{we're in the car} then this was her last day of school. At the bus stop it was all so bad that I'd have forgotten not to say goodbye " Hello, It seems I've had the worst day of my life. Please remember to check on Jane " " Excuse me child " I looked at her puzzled " oh my god " A strange feeling came over me and I knew I didn't have time for that in the moment. " K- Come on Jane " Suddenly I became stiff and lost my footing. I fell onto the curb and immediately collapsed onto the curb again. " What the HELL" followed by everything I missed out on. I knew my grandmother was trying to get me to go to the movies.
 \end{minipage} }
   \caption{Generation from GPT-117 fine-tuned without \eop/\sep.} 
\end{figure}

\begin{figure}[!htb]
\centering \small
\fbox{
\begin{minipage}{.95\textwidth}
My heart was racing; I felt \textbf{something otherworldly being in my car}. " Hello again! Who is this? " I was already bracing for another \textbf{crash} at work.\\
" Shhh shhh " the object before me suddenly let go. I tried turning it on to get some air. Nope, this thing didn't change its positioning, it never got to anything. My arm feels like a pressure cooker, its chest feels like a heater. In all that time I never had expected to see anything like this happening.
My legs were numb, and my back was numb. I was so tired. I was like a boat out of water and it seemed like it took all my energy to get the \textbf{gas pedal off}. " I swear to god please don't hurt me! " I screamed, and I was feeling for my hands like claws were opening my right leg.\\
Suddenly the skull popped open... \textbf{my car began to fly}..no where else to go except to look on the horizon and never see it again. I had been driving home in rage and frustration. \textbf{I need to get out of here}, I'm literally the black dog everywhere. "\textbf{ What the hell happened}? " I was about to ask out of fear of being right next to someone when a screeching fell from above. I slowly turned my head towards the heavens. \textbf{Something was definitely different in the sky.}\\
" Can you hear me? " I shouted, though I was standing a full 6 feet tall. I was so sure. * What is going on? * I wanted to yell out, to pretend that I didn't hear someone. " Help me! Please! " What! No. Please. Please help me! " I must have just tried to reach for my hand or something.\\
It landed in the grassland and started shaking. My chest felt heavier and it was hardly noticeable on the ground. It shook harder, and now my eyes were fixated on that star. " Fuuuuuk, Fuuuuuuuuuk... " *BANG*, the Earth shook. I was spinning and was floating. It was so bright I looked down at my arms, and \textbf{I felt like we were floating for what seemed like an eternity}. I was breathing like I was being ripped apart.\\
Then I felt something was dragging me. I don't know what it was, but it was drifting closer and closer and more and more until I felt as though I was about to fall and I could hear what could only be a small nervous chuckle escaping my lips. " God this is so weird! "\\
" What're you doing? "\\
" I'm.... In a strange alternate world.\textbf{ I'm falling from this sky}, touching a comet... "\\
" Hell no! This is \textbf{just a dream}... I'm falling from this sky and it's going to explode and everything is going to be fine until I touch a comet again. Then... this is it. All of this is right now and I can't see what's going on but I can feel it. I can feel it being lifted off of me and all of this knowledge and knowledge... "\\
*Brilliant* What?! It was the best dream I had ever had! I could definitely feel that tiny little planet being pulled up to the surface by whatever ship it was in. God, could I ever know what will happen to it like that? But what would happen to me if I touched one such comet! Even being floating in space with nothing to do?\\
I had finally had enough! " I'll get you out of here! I'll let you out... ( Once I get to the comet, I can feel the universe get pulling me to another dimension. ) I'll let you out! " I took off my sunglasses and put them on, and there was no way I wouldn't be in that crowd again.\\
After that nightmare I decided that I wasn't going to make it out of here alive. That I couldn't \textbf{ride a car }that would almost definitely end in killing someone. Whatever! I'll be back, that's all I need.\\
\textbf{I'll be free of this reality, *that dream* and open to some other universe!}
 \end{minipage} }
   \caption{Generation from GPT-117 fine-tuned with \nl~as \sep.} 
\end{figure}
\clearpage
\begin{figure}[!htb]
\centering \small
\fbox{
\begin{minipage}{.95\textwidth}
Here we go again, I'm trying to pick up my pace. My hands are dragging along the overgrowth and I hardly feel a thing. The sunlight beating down on me still yet dulling my vision. My hand suddenly feels like it's in a basketball as I pick myself up off the ground. \\
 "Hey Wendy... You okay? " \\
 " Just trying to get some fresh air on my lungs " \\
 " Can you get me one of my of my stones to drink. I mean, \textbf{there is no use in driving} " \\
 " Sure... " I say looking around, taking up a new position on a nearby skylight. While searching, one on a hill that seems pretty rounded in any way, two other ones are out of order, and I take notice of them trying to move out of the way. \\
 Not even half an hour passes... I can feel them roll in and out of the path as I decide it's time to head out for the day. No, I don't give one. \\
 " Would you like some fresh air for my shoulder? " \\
 " How about that new Sonic X that's been around for the past couple years? Now as soon as I get off it, it can take me out of the sun. So just give me a moment of peace and rest " \\
 I reach for my rock, still still clutching at my leg with my shoe. Yet as fast as I left it, it's trapped by my arm. I'm powerless to do anything... until I hear something coming down from the trees. " STOP! " I yell as I try to dodge it in a fast spiral. Before I can react, it's shoved right at me and\textbf{ I fall to the ground}. The sky is dark, smog filling the sky. Already I'm blacking out, the\textbf{ backlight on my car keeping me firmly in darkness.} \\
 A crisp wind whipping about me, I grab my blanket from my chair and prepare to throw it at anything that could have managed to keep me with me. Bouncing out of my chair, I continue down the path where the road once was. \\
 The wind is beginning to get stronger. More thunderstorms begin breaking out, as well as additional thunder. My turn comes up and the wind picks up. As soon as I can see it, it's nowhere to be seen. I'm only about 10 minutes away from the road, standing in the middle of the road. I hear \textbf{a voice screaming from my car}. A tall man in fatigues looks at me and \textbf{looks at my car}. " Damn... \textbf{I was driving}... " he says, before sprinting from my car and grabbing his wallet. He gives me a look of disgust, as if the only thing worse than avoiding the \textbf{highway} was choosing between two other men. \\
 I ask him what was going on, and he smiles gently. " You think I'm lucky to get in, huh? \textbf{I really shouldn't be riding a car just yet}, you know. But I'm glad you're here! So if you don't mind if I drive me, I have a few things on my mind. " \\
 " Alright, fine, whatever. Go, fasten the seat belt, you can't come back here any other way. Are you sure you're just going to excuse me, though?" \\
 That was his last expression, before he limped away like a glutton.\\ 
 \textbf{This is the end of my first attempt at writing nothing! Any thoughts of how to improve upon it?}
 \end{minipage} }
   \caption{Generation from GPT-117 fine-tuned with DIY \eop.} 
\end{figure}